\documentclass[letterpaper]{article} 
\usepackage{aaai21}  
\usepackage{times}  
\usepackage{helvet} 
\usepackage{courier}  
\usepackage[hyphens]{url}  
\usepackage{graphicx} 
\usepackage{graphicx}  
\usepackage[numbers]{natbib}  
\usepackage{caption} 
\usepackage{amsmath}
\usepackage{hyperref}
\usepackage[switch]{lineno} 

\title{Consumer Behaviour in Retail: Next Logical Purchase using Deep Neural Network}

\author{

    Ankur Verma
}
\affiliations{

    Samya.ai \\
    ankur.verma@samya.ai
}

\begin{document}
\maketitle

\begin{abstract}
Predicting future consumer behaviour is one of the most challenging problems for large scale retail firms. 
Accurate prediction of consumer purchase pattern enables better inventory planning and
efficient personalized marketing strategies. Optimal inventory planning helps minimise instances of Out-of-stock/ Excess Inventory 
and, smart Personalized marketing strategy ensures smooth and delightful shopping experience.
Consumer purchase prediction problem has generally been addressed by ML researchers in conventional manners, either 
through recommender systems or traditional ML approaches. Such modelling approaches do not generalise well 
in predicting consumer purchase pattern. In this paper, we present our study of consumer purchase behaviour, wherein,
we establish a data-driven framework to predict whether a consumer is going to purchase an item 
within a certain time frame using e-commerce retail data. To model this relationship, we create a sequential time-series data
for all relevant consumer-item combinations. We then build generalized non-linear models by generating features
at the intersection of consumer, item, and time. We demonstrate robust performance by experimenting with different neural 
network architectures, ML models, and their combinations. We present the results of 60 modelling experiments with 
varying Hyperparameters along with Stacked Generalization ensemble \cite{wolpert1992stacked} and
F\textsubscript{1}-Maximization framework. We then present the benefits that neural network architectures like Multi 
Layer Perceptron, Long Short Term Memory (LSTM), Temporal Convolutional Networks (TCN) \cite{lea2016temporal} and 
TCN-LSTM \cite{karim2017lstm} bring over ML models like Xgboost \cite{chen2016xgboost} and RandomForest. 

\end{abstract}
\section{Introduction}
Consumer behaviour insights have always been one of the key business drivers for retail, given
fast changing consumer needs. Existing trend, competitor pricing, item reviews, sales and marketing are some of the 
key factors driving today's consumer world in retail. While very little information is available
on future variablities of the above factors, retailers do have large volumes of historical transactional data. Past study 
\cite{choudhury2019machine} has shown that retailers use conventional techniques with available data to model consumer purchase. 
While these help in estimating purchase pattern for loyal consumers and high selling items with reasonable accuracy, they 
do not perform well for the long tail. Since multiple parameters interact non-linearly to define consumer purchase pattern,
traditional models are not sufficient to achieve high accuracy across thousands to millions of consumers.

Most retail/e-retail brands, plan their short term inventory (2-4 weeks ahead)  based on consumer 
purchase pattern. Also, certain sales and marketing strategies like Offer Personalization and personalized item
recommendations are made leveraging results of consumer purchase predictions for the near future.
Given that every demand planner works on a narrow segment of item portfolio, there is a high 
variability in choices that different planners recommend. Additionally, the demand planners might not get enough opportunities 
to discuss their views and insights over their recommendations. Hence, subtle effects like cannibalization
\cite{shah2007retailer}, and item-affinity remain unaccounted for. Such inefficiencies lead to a gap between consumer needs 
and item availability, resulting in the loss of business opportunities in terms of consumer churn, and out-of-stock
and excess inventory.

Our paper makes the following contributions -
\begin{itemize}
\item We study and present the usefulness of applying various deep learning architectures along with tree based machine 
learning algorithms to predict the next logical item purchase at consumer level.
\item We present the performance of individual models with varying hyperparameter configurations.
\item We implement stacked generalization framework \cite{wolpert1992stacked} as an ensemble method where a new model learns 
to combine the predictions from multiple existing models.
\item We design and implement F\textsubscript{1}-maximization algorithm which optimises for purchase probability cut-off 
at consumer level.
\end{itemize}

\section{Related Work}
\label{sec:relatedwork}
In the past few years, usefulness of various machine learning methods for predicting consumer purchase pattern have been 
analyzed in the academia field and few of them are often used by ML practitioners. In most cases many of those approaches are 
based on extracting consumer's latent characteristics from its past purchase behavior and applying statistical and  
ML based formulations \cite{fader2009probability, choudhury2019machine}. 
Some previous studies have analyzed the use of random forest and Xgboost techniques to predict 
consumer retention, where past consumer behavior was used as potential explanatory variable 
for modelling such patterns. In one such study \cite{martinez2020machine}, the authors develop a model for predicting whether a 
consumer performs a purchase in prescribed future time frame based on historical purchase information such as the number
of transactions, time of the last transaction, and the relative change in total spending of a consumer. 
They found gradient boosting to perform best over test data. We propose neural network architectures with entity embeddings
\cite{guo2016entity} which outperform the gradient boosting type of models like Xgboost \cite{chen2016xgboost}. 

From Neural Network architectures perspective,
close to our work is Deep Neural Network Ensembles for Time Series Classification \cite{fawaz2019deep}. 
In this paper, authors show how an ensemble of multiple Convolutional Neural Networks can improve upon the 
state-of-the-art performance of individual neural networks. They use 6 deep learning classifiers 
including Multi Layer Perceptron, Fully Convolutional Neural Network, Residual Network, 
Encoder \cite{serra2018towards}, Multi-Channels Deep Convolutional Neural Networks \cite{zheng2014time} and 
Time Convolutional Neural Network \cite{zhao2017convolutional}. The first three were originally proposed in \cite{wang2017time}.
We propose the application of such architectures in the consumer choice world and apply the concept of entity embeddings 
\cite{guo2016entity} along with neural network architectures
like Multi Layer Perceptron, Long Short Term Memory (LSTM), Temporal Convolutional Networks (TCN) \cite{lea2016temporal} and 
TCN-LSTM \cite{karim2017lstm}.

\section{Methodology}
\label{sec:methodology}
We treat each relevant consumer-item as an individual object and shape them into weekly time series data
based on historical transactions. In this setup, target value at each time step (week) takes a binary input, 1/0 
(purchased/non purchased). \emph{Relevancy} of the consumer-item is defined by items transacted by consumer during training 
time window. \emph{Positive samples} (purchased/1) are weeks where consumer did transact for an item, whereas 
\emph{Negative samples} (non purchased/0) are the weeks where the consumer did not buy that item.
We apply sliding windows testing routine for generating
out of time results. The time series is split into 4 parts - train, validation, 
test1, and test2 as shown in Table \ref{tab:datasplit}. All our models are built in a multi-object 
fashion, which allows the gradient movement to happen across all consumer-item combinations split in batches. This enables 
cross-learning to happen across consumers/items. We then perform Feature Engineering over data splits to generate
modelling features. Below are some of the feature groups we perform our experiments with:
\begin{itemize}
\item {\bf Datetime:} We use transactional metrics at various temporal cuts like week, month, etc.
Datetime related features capturing seasonality and trend are also generated.
\item {\bf Consumer-Item Profile:} We use transactional metrics at different granularities like consumer, item,
consumer-item, department and aisle. We also create features like Time since first order, 
Time since last order, time gap between orders, Reorder rates, Reorder frequency, 
Streak - user purchased the item in a row, Average position in the cart, Total number of orders.
\item {\bf Consumer-Item-Time Profile:} We use transactional metrics at the intersection of consumer, item and time.
We generate interactions capturing consumer behaviour towards items at a given time.
\item {\bf Lagged Offsets:} We use statistical rolling operations like mean, median, quantiles, variance, 
kurtosis and skewness over temporal regressors for different lag periods to generate offsets.
\end{itemize}
The model we need to build, thus, should learn to identify similarly behaving time series across latent
parameters, and take into account consumer and item variations in comparing different time series. A row 
in time series is represented by
  \begin{equation}
    \begin{array}{l}
      y\textsubscript{cit}  = f(i\textsubscript{t}, c\textsubscript{t},..,c\textsubscript{t-n}, ic\textsubscript{t}
      ,..,ic\textsubscript{t-n}, d\textsubscript{t},..,d\textsubscript{t-n})
    \end{array}
    \label{eqn:fx}
  \end{equation}
where y\textsubscript{cit} is purchase prediction for consumer 'c' for item ’i’ at time ’t’. 
i\textsubscript{t} denotes attributes of item ’i’ like category, department, brand, color, size, etc at time 't'. 
c\textsubscript{t} denotes attributes of consumer 'c' like age, sex and transactional attributes at time 't'. 
ic\textsubscript{t} denotes transactional attributes of consumer 'c'  towards item 'i' at time 't'. 
d\textsubscript{t} is derived from datetime to capture trend and seasonality at time 't'. 
'n' is the number of time lags.
\begin{table}[t]
\caption{Modelling data splits}
\vspace{0.1 in}
\centering
\resizebox{3.3in}{!}
{%
\begin{tabular}{|c|c|c|c|}
\hline
{\bf Data Split} & {\bf Specifications} & {\bf Consumer-Item combinations} & {\bf Max Time-Series length} \\  
\hline\hline
Train  		&  Model training &  50,872 &  46 weeks \\ \hline
Validation	  		&  HyperParameter Optimization &  50,888 &  2 weeks \\ \hline
Test1  		&  Stacked Generalization, F\textsubscript{1}-Maximization & 50,899 &  2 weeks\\ \hline
Test2	  		&  Reporting Accuracy Metrics & 50,910 &  2 weeks\\
\hline
\end{tabular}
}
\label{tab:datasplit}
\end{table}
\subsection{Loss Function}
Since we are solving Binary Classification problem, we believe that Binary Cross-Entropy should be the most appropriate 
loss function for training the models. We use the below formula to calculate Binary Cross-Entropy:
  \begin{equation}
      \begin{array}{l}
        H\textsubscript{p} = - \frac{1}{N}$$\sum_{i=1}^{N}y.log(p(y))+
        (1- y).log(1-p(y))
      \end{array}
    \label{eqn:logloss}
  \end{equation}
here H\textsubscript{p} represents computed loss, y is the target value (label), and p(y) 
is the predicted probability against the target. The BCELoss takes non-negative values. We can infer 
from Equation \ref{eqn:logloss} that Lower the BCELoss, better the Accuracy.
\subsection{Model Architectures}
\begin{figure}[t]
    \centering 
    \caption{Data classification for DNN Architectures} 
    \includegraphics[width=3.3in]{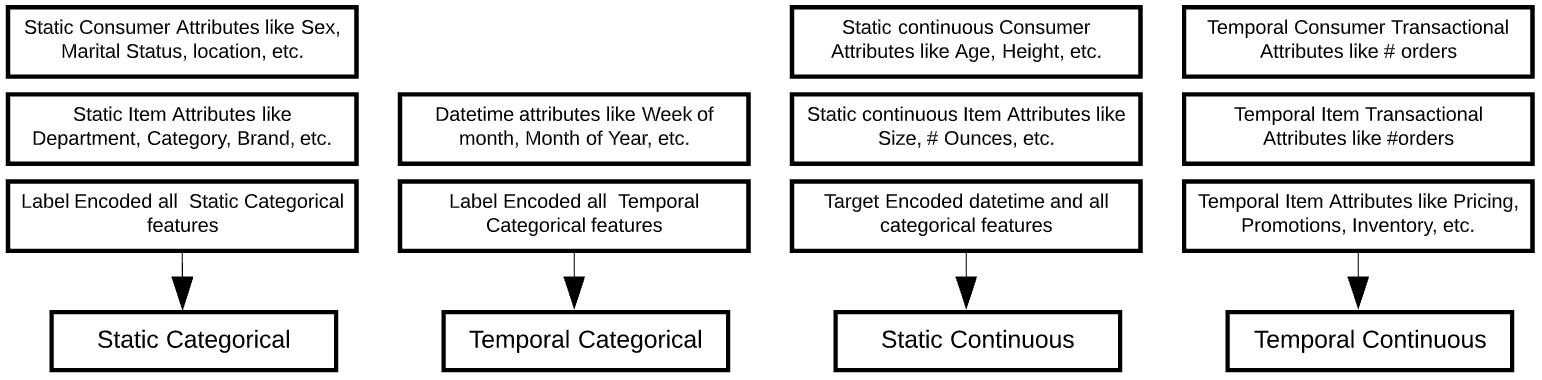} 
    \label{fig:dnndata} 
  \end{figure}

\begin{figure}[t]
    \centering 
    \caption{Multi Layer Perceptron (MLP)} 
    \includegraphics[width=3.3in]{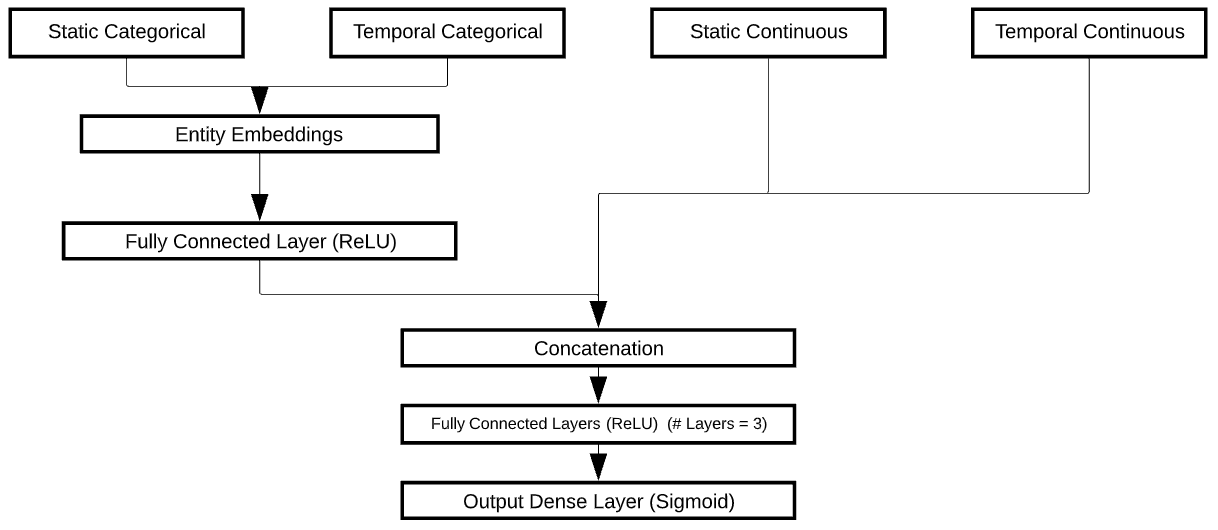} 
    \label{fig:MLP} 
  \end{figure}

\begin{figure}[t]
    \centering 
    \caption{Long Short Term Memory (LSTM)} 
    \includegraphics[width=3.3in]{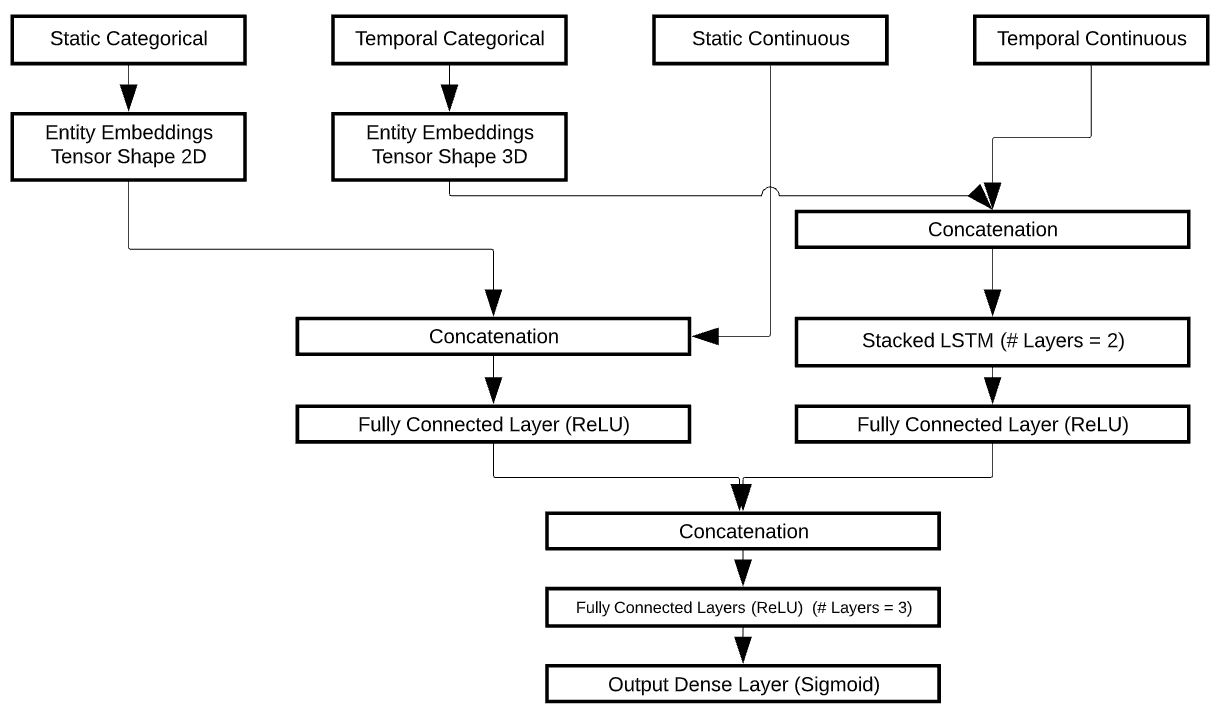} 
    \label{fig:LSTM} 
  \end{figure}

\begin{figure}[t]
    \centering 
    \caption{Temporal Convolutional Network (TCN)} 
    \includegraphics[width=3.3in]{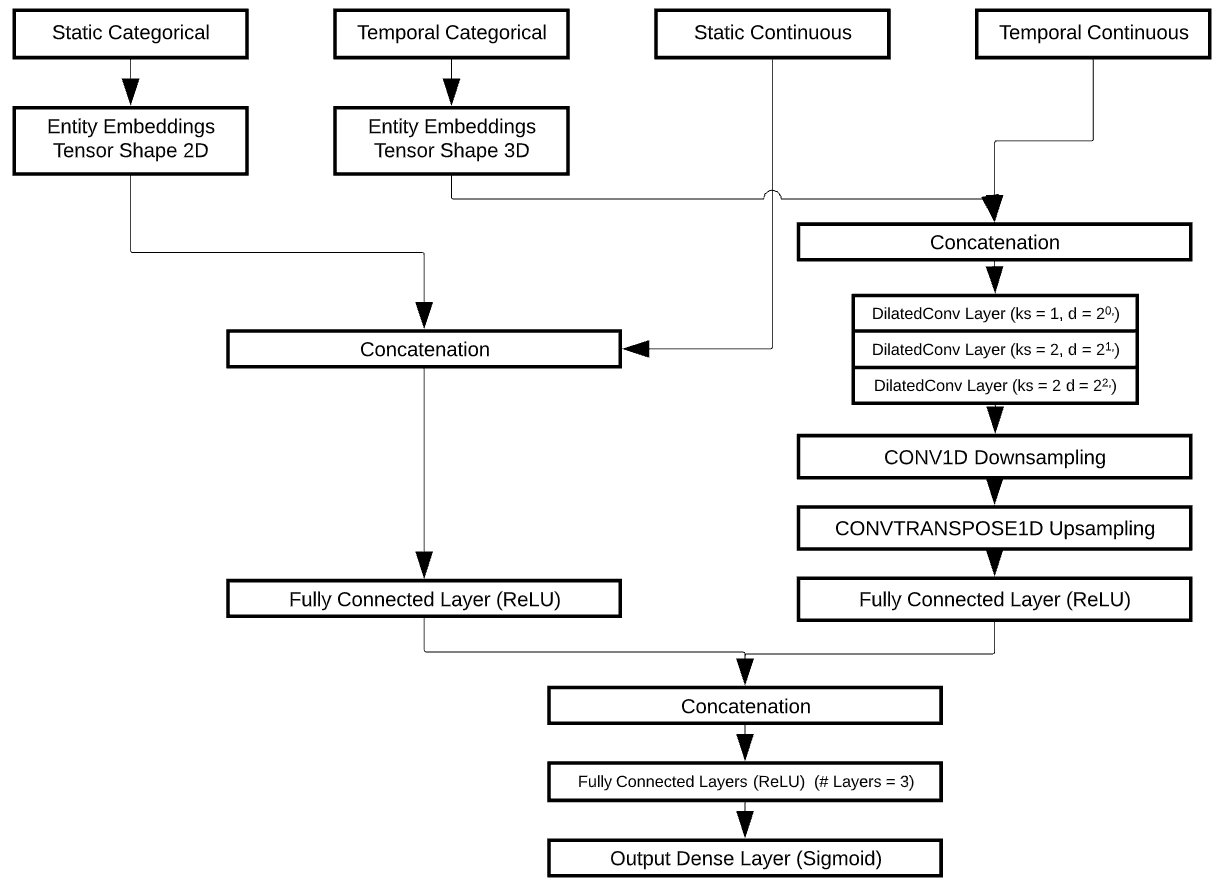} 
    \label{fig:TCN} 
  \end{figure}

\begin{figure}[t]
    \centering 
    \caption{TCN-LSTM} 
    \includegraphics[width=3.3in]{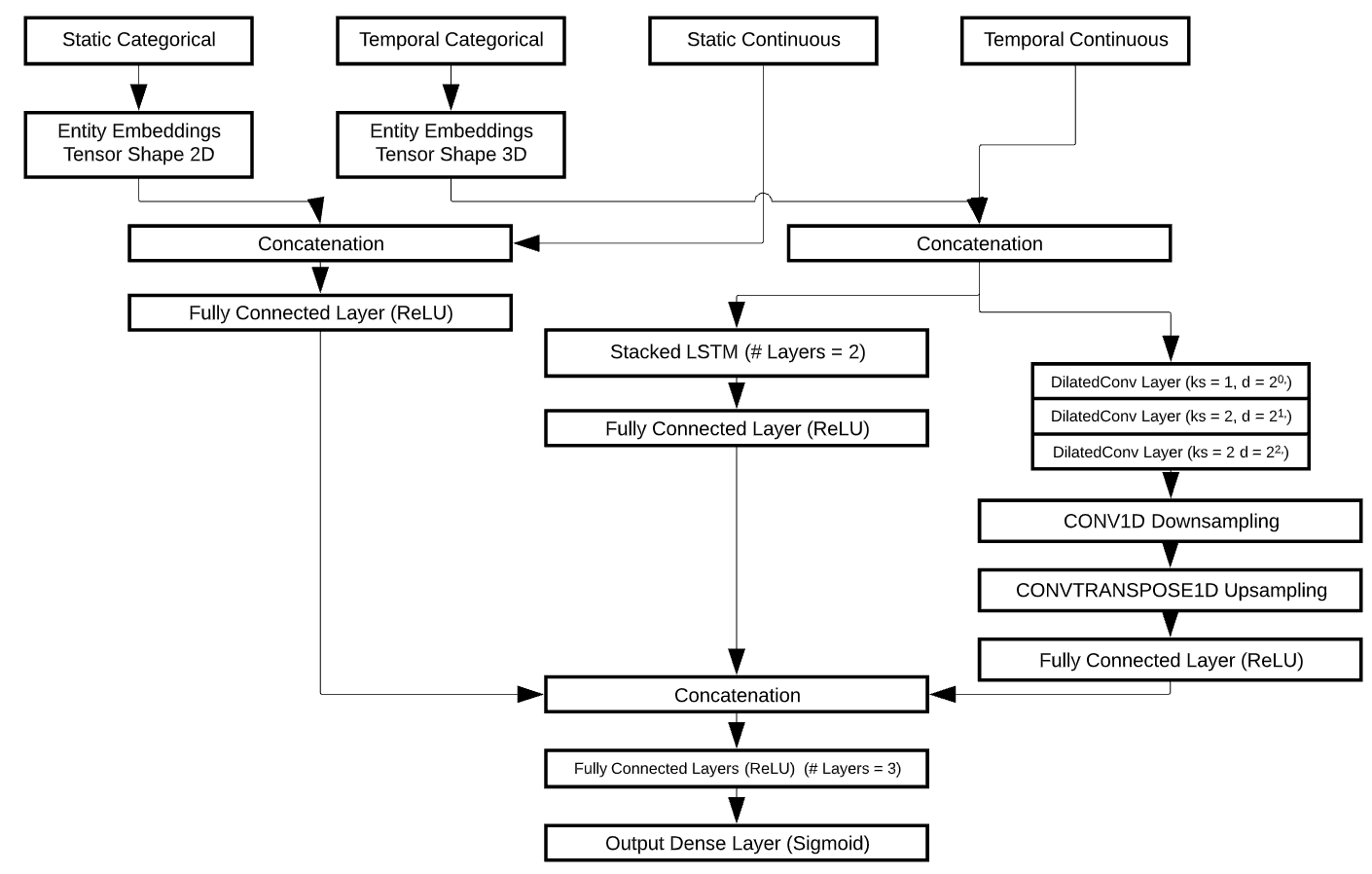} 
    \label{fig:TCN-LSTM} 
  \end{figure}

As mentioned earlier in this section, traditional machine learning models are not really a suitable choice for modelling \emph{f} 
(Equation \ref{eqn:fx}) due to non-linear interaction between the features. Hence, we work with
tree based models like RandomForest, Xgboost \cite{chen2016xgboost} 
to Deep learning models like Multi Layer Perceptron (MLP), Long Short 
Term Memory (LSTM) and Temporal Convolutional Networks (TCN). Architectures of MLP, LSTM, TCN \cite{lea2016temporal} 
and TCN-LSTM \cite{karim2017lstm} models are shown in Figure \ref{fig:MLP}, Figure \ref{fig:LSTM}, Figure \ref{fig:TCN}
and Figure \ref{fig:TCN-LSTM} respectively. We briefly describe the architectures below.
\begin{itemize}
\item {\bf Entity Embeddings + Multi Layer Perceptron:} MLP (Figure \ref{fig:MLP}) is the simplest form of deep neural networks 
and was originally proposed in \cite{wang2017time}. The architecture contains 
three hidden layers fully connected to the output of its previous layer. The final layer 
is the sigmoid layer which generates the probability. One disadvantage of MLP is that since the input time 
series is fully connected to the first hidden layer, the temporal information in a time series is lost \cite{fawaz2019deep1}.
\item {\bf Entity Embeddings + Long Short Term Memory:} LSTM (Figure \ref{fig:LSTM}) is an
architecture comprising of 2 LSTM layers combined with entity embeddings. This combination flows into 
3 fully connected ReLU based layers yielding to dense layer which has sigmoid activation.
\item {\bf Entity Embeddings + Temporal Convolutional Network:} TCN (Figure \ref{fig:TCN}), originally
proposed in \cite{lea2016temporal} , is considered a competitive architecture yielding the best results when evaluated over 
our experimental dataset. This network comprises of 3 dilated Convolutional networks combined with entity embeddings.
Similar to LSTM, this architecture, after convolving and concatenating flows into 
3 fully connected ReLU based layers yielding to dense layer which has sigmoid activation.
\item {\bf Entity Embeddings + Long Short Term Memory-Temporal Convolutional Network:} TCN-LSTM (Figure \ref{fig:TCN-LSTM})
inherits the properties of LSTM and TCN in a fully connected network.
\end{itemize}
From data classification, Figure \ref{fig:dnndata}, we can see that data was sub-divided 4 groups:
\begin{itemize}
\item {\bf Static Categorical:} These are categorical features that do not vary with time. This includes consumer
attributes like sex, marital status and location along with different item attributes like category, department and brand.
\item {\bf Temporal Categorical:} These are categorical features that vary with time. It includes all the datetime 
related features like week, month of year, etc.
\item {\bf Static Continuous:} These features are static but continuous. This includes certain consumer attributes like
age and weight, item attributes like size, and certain derived features like target encoded features.
\item {\bf Temporal Continuous:} These are time varying continuous features. All consumer and item related
traditional attributes like number of orders, add to cart order, etc. falls under this bucket.
\end{itemize}
As mentioned earlier in this section, in all the above described neural network architectures, we learn the embeddings \cite{guo2016entity} of the 
categorical features during training phase. We embed these attributes in order to compress their representations 
while preserving salient features, and capture mutual similarities and differences.

The consumer purchase pattern has huge variation in terms of time of purchase (weekday/weekends), 
cadence of purchase (days to months), purchased item types (dairy/meat/grocery/apparels/etc.)
and brand loyalty (tendency to substitute items). Given such huge variations, it becomes imperative 
to cross learn consumer behaviour from similar consumer groups. To learn such relationships its 
important to capture non-linear relationship between target and regressors at various levels.
Tree based and Deep learning models are chosen for their ability to model feature interactions (even if transient in time), 
and their ability to capture non-linearity well.

Tree based models and MLP training uses lagged values of time varying regressors
to capture temporal dependencies. LSTM, TCN and TCN-LSTM models are trained using entire life-cycle data of a 
time series (Consumer-Item) in sequential manner. Details around dataset and features are explained in section \ref{sec:eval}. 

\subsection{Hyperparameter Tuning}
Hyper-parameters of tree based models are optimized
using Bayesian Hyperparameter Optimization Technique, Hyperopt \cite{bergstra2013hyperopt}. 
For Deep learning, we use documented best practices along with our experimental results to
choose model hyperparameters. Hyperparameter Optimization is performed over validation dataset. 
We list some of the hyperparameters along with the values we tune for Deep learning models.
  \begin{itemize}
    \item {\bf Optimizer Parameters:} RMSProp \cite{bengio2015rmsprop} and Adam are used as different trial configurations. 
    The learning rate is experimentally tuned to 1e-3. We also have weight decay of 1e-5 which helps a bit in model Regularization.
    \item {\bf Scheduler Parameters:} CyclicLR \cite{smith2017cyclical} and ReduceLROnPlateau \cite{zaheer2018adaptive} 
    Learning rates are used as different trial configurations.
    we use 1e-3 as max lr and 1e-6 as base lr for cyclical learning rate along with the step size being the function of
    length of train loader. ReduceLROnPlateau is tuned at 1e-6 as min lr.
    \item {\bf SWA:} Stochastic Weight Averaging (SWA) \cite{izmailov2018averaging} is used to improve generalization across Deep Learning
    models. SWA performs an equal average of the weights traversed by SGD with a modified learning rate schedule. We use 
    1e-3 as SWA learning rate.
    \item {\bf Parameter Average:} This is a method to average the neural network parameters of n best model checkpoints 
    post training, weighted by validation loss of respective checkpoints. The resulting model generalizes better than those 
    with a single best checkpoint model for an unseen data. 
  \end{itemize}
Apart from the above parameters we also iterate to tune network parameters like number of epochs, batch size, 
number of Fully Connected Layers, number of LSTM layers, convnet parameters (kernel size, dilations, padding)
and embedding sizes for the categorical features. Binary Cross-Entropy \ref{eqn:logloss} is used as loss 
function for all the models. For sequence model \cite{sutskever2014sequence}, we also iterate with
Dropout \cite{hinton2012improving} and BatchNorm \cite{santurkar2018does} within networks.
Hyperparameters used for Machine learning models with Hyperopt \cite{bergstra2013hyperopt} are :
  \begin{itemize}
    \item {\bf Learning Rate:} Range set to vary between 1e-2 to 5e-1. 
    \item {\bf Max Depth:} Range set from 2 to 12 at step of 1.
  \end{itemize}
Apart from above hyperparameters, regularization parameters like Reg Lambda and Min Sample Leaf are also optimized using Hyperopt.

Deep learning models are built using deep learning framework
PyTorch \cite{paszke2017automatic}, and are trained on GCP instance containing 6 CPUs and a single GPU. 
Scikit-learn \cite{pedregosa2011scikit} is used for Tree
based models like RandomForest and Xgboost \cite{chen2016xgboost}. For Neural Network Architectures, 
we save model weights of the best checkpoint, so that we may access the learned entity embeddings and other
model weights if required. As described in Table \ref{tab:modelparams}, we build a total of 60 models, 
12 different configurations for each of 4 Deep Learning models and 6 best trials from Hyperopt \cite{bergstra2013hyperopt} 
for each of 2 Machine Learning models.
\subsection{Stacked Generalization Ensemble}
Stacked generalization or Stacking \cite{wolpert1992stacked} is an ensemble method where a new model learns how to best 
combine the predictions from multiple existing models. This is achieved by training an entirely new model using 
contributions from each submodel. 
We use weighted K-best model as Stacker for combining k submodels (candidates) out of total 60 
trained models. We iterate with different values of k ranging from 3 to 25 as presented in Table \ref{tab:stacking} . 
test1 BCELoss of submodels is used for weight initialization for the stacker models. For learning optimal weights of submodels, 
we minimise test1 BCELoss of the stacker model using gradient descent \cite{ruder2016overview},
stacking function can be described as:
  \begin{equation}
    \begin{array}{l}
      y\textsubscript{cit} =\sum_{j=1}^{k}w\textsubscript{j} \times p\textsubscript{cit\textsubscript{j}}
    \end{array}
    \label{eqn:stack}
  \end{equation}
where y\textsubscript{cit} is the stacked probability for consumer 'c' for item ’i’ at time ’t’.
k represents the number of candidates shortlisted for stacking, p\textsubscript{cit\textsubscript{j}}
represents the prediction probability for consumer 'c' for item ’i’ at time ’t’ by j\textsuperscript{th} submodel.
w\textsubscript{j} is the weight for j\textsuperscript{th} submodel.

\subsection{F\textsubscript{1}-Maximization}
Post stacking, we optimize for purchase probability threshold based on
probability distribution at a consumer level using F\textsubscript{1}-Maximization.
This enables optimal thresholding of consumer level probabilities to  maximize F\textsubscript{1} measure \cite{lipton2014optimal}.
To illustrate the above, let us say we generated purchase probabilities for 
'n' items out of 'b' actually purchased items by consumer 'c'. Now, let us visualize the actuals (\ref{eqn:A}) 
and predictions (\ref{eqn:P})  of 'n' predicted items for consumer 'c'.
  \begin{equation}
    \begin{array}{l}
      A\textsubscript{c} = [a\textsubscript{1}, a\textsubscript{2}, .., a\textsubscript{n}] 
       \; \forall \; a\textsubscript{j} \in \; $\{0,1\}$
    \end{array}
    \label{eqn:A}
  \end{equation}
  \begin{equation}
    \begin{array}{l}
      P\textsubscript{c} = [p\textsubscript{1}, p\textsubscript{2}, .., p\textsubscript{n}]
      \; \forall \; p\textsubscript{j} \in \; [0,1]
    \end{array}
    \label{eqn:P}
  \end{equation}
A\textsubscript{c} represents the actuals for consumer 'c', with a\textsubscript{j} being 1/0 
(purchased/non purchased). P\textsubscript{c} represents the predictions 
for consumer 'c' for the respective item, with p\textsubscript{j} being probability value. 
'n' represents total items for which the model generated purchase probabilities for consumer 'c'.
Now we apply Decision rule D() which converts probabilities to binary predictions, as described below 
in Equation \ref{eqn:Decision}.
  \begin{equation}
    \begin{array}{l}
      D(Pr\textsubscript{c}) : P\textsubscript{c}\textsuperscript{1 x n}
      \to P\textsuperscript{'}\textsubscript{c}\textsuperscript{1 x n}
      \;\; : p\textsuperscript{'}\textsubscript{j} = 
        \begin{cases}
          1 & p\textsubscript{j} \geq Pr\textsubscript{c} \\
          0 & \text{Otherwise}
        \end{cases}
    \end{array}
    \label{eqn:Decision}
  \end{equation}
  \begin{equation}
    \begin{array}{l}
      P\textsuperscript{'}\textsubscript{c} = [p\textsuperscript{'}\textsubscript{1}, 
      p\textsuperscript{'}\textsubscript{2}, .., p\textsuperscript{'}\textsubscript{n}]\; 
      \forall \; p\textsuperscript{'}\textsubscript{j} \in \; $\{0,1\}$
    \end{array}
    \label{eqn:Pdash}
  \end{equation}
  \begin{equation}
    \begin{array}{l}
      k =\sum_{i=1}^{n}p\textsuperscript{'}\textsubscript{i}
    \end{array}
    \label{eqn:Pdash}
  \end{equation}
Pr\textsubscript{c} is the probability cut-off to be optimized for maximizing F\textsubscript{1} measure \cite{lipton2014optimal} 
for consumer 'c'. Decision rule D() converts probabilities P\textsubscript{c} to binary predictions 
P\textsuperscript{'}\textsubscript{c} such that if p\textsubscript{j} is less than 
Pr\textsubscript{c} then p\textsuperscript{'}\textsubscript{j} equals 0, otherwise 1.
'k' is the sum of predictions generated post applying Decision rule D(). Now we solve for F\textsubscript{1} measure
using equations and formulae described below.
  \begin{equation}
    \begin{array}{l}
      V\textsubscript{Pr\textsubscript{c}} = 
      P\textsuperscript{'}\textsubscript{c}
      \;\times\; A\textsubscript{c}\textsuperscript{T}
      \;
      \Rightarrow	
      \left( \begin{array}{ccc}
      p\textsuperscript{'}\textsubscript{1} & .. & 
      p\textsuperscript{'}\textsubscript{n}
      \end{array} \right)
      \times
      \left( \begin{array}{ccc}
      a\textsubscript{1} \\
      .. \\
      a\textsubscript{n} \\
      \end{array} \right)
    \end{array}
    \label{eqn:probability}
  \end{equation}
  \begin{equation}
    \begin{array}{l}
      Precision\textsubscript{c}= \frac{V\textsubscript{Pr\textsubscript{c}}} {k}
      \;\;\;\;\;and\;\;\;\;
      Recall\textsubscript{c}= \frac{V\textsubscript{Pr\textsubscript{c}}} {b}
    \end{array}
    \label{eqn:F1}
  \end{equation}
  \begin{equation}
    \begin{array}{l}
      F\textsubscript{1\textsubscript{c}} = \frac{2 \times Precision\textsubscript{c} 
      \times Recall\textsubscript{c}} 
      {Precision\textsubscript{c} + Recall\textsubscript{c}}
      \;
      \;\;\;\;\Rightarrow	\;\;\;\;
      2 * 
      \frac{
        V\textsubscript{Pr\textsubscript{c}}
      }
      {
        k + b
      }
    \end{array}
    \label{eqn:Optimizer}
  \end{equation}
V\textsubscript{Pr\textsubscript{c}} represents the number of items with purchase 
probabilities greater than Pr\textsubscript{c} which were actually purchased (True Positives). 
As can be seen, Formulae \ref{eqn:F1} and \ref{eqn:Optimizer} are used to calculate Precision, Recall and 
F\textsubscript{1}-score for consumer 'c'. 

  \begin{equation}
    \max_{V\textsubscript{Pr\textsubscript{c}}} \;\;\;\; 2 * \frac{ V\textsubscript{Pr\textsubscript{c}}}{k + b}
    \;\;\;\;,\;\;\;\;  \text{subject to: } \;\;\;\;  Pr\textsubscript{c}  \in \; [0,1]
    \label{eq:constraint}
  \end{equation}
Equation \ref{eq:constraint} represents the optimization function we solve to generate purchase predictions (1/0) for each consumer.
\begin{center}
\begin{table*}[!t]
\caption{Model Specifications}
\centering
\resizebox{\textwidth}{!}{\begin{tabular}{|r|l|r|r|}
  \hline
 {\bf Model Type} & {\bf Trials} & {\bf Model HyperParameters} & {\bf Loss Functions}\\ 
 \hline\hline
MLP	  		&  12 & Optimizer, Scheduler, SWA, Parameter Averaging, Feature Groups, FC Layers & BCELoss\\ \hline
LSTM  		& 12 & Optimizer, Scheduler, SWA, Parameter Averaging, Feature Groups, FC Layers, LSTM Layers & BCELoss \\ \hline
TCN			& 12	& Optimizer, Scheduler, SWA, Parameter Averaging, Feature Groups, FC Layers, Convolution Parameters  & BCELoss\\ \hline
TCN-LSTM 		& 12	& Optimizer, Scheduler, SWA, Parameter Averaging, Feature Groups, FC Layers, LSTM, Convolution Parameters  & BCELoss\\ \hline
Xgboost 		& 6	& Learning rate, Tree Depth, Regularization parameters  & BCELoss\\ \hline
RandomForest 		& 6	& Tree Depth, Evaluation Metrics, Regularization parameters &  BCELoss\\
   \hline
\end{tabular}}
\label{tab:modelparams}
\end{table*} 
\end{center}

\section{Experiments and Results}
\label{sec:eval}
\begin{figure*}[!t]
    \centering 
    \caption{Sample Dataset} 
    \includegraphics[width=6.6in]{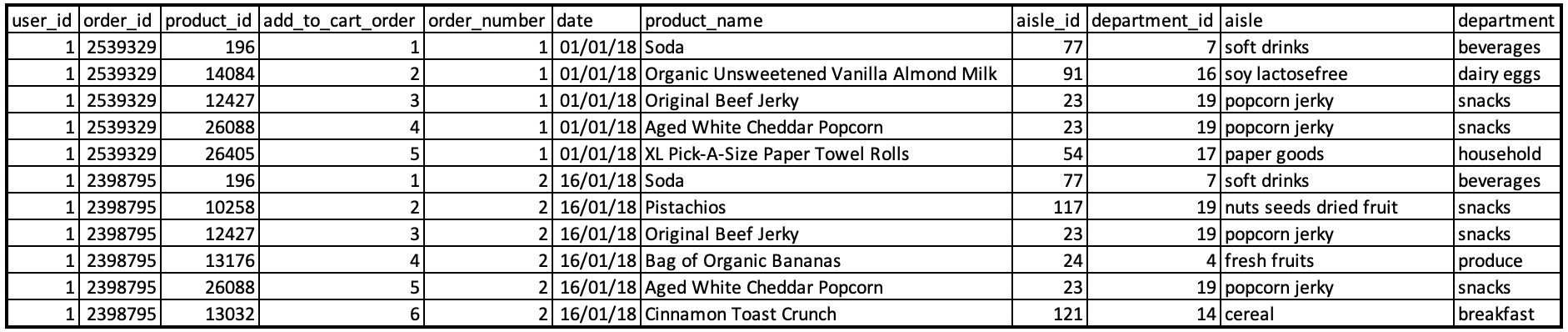} 
    \label{fig:sampledata} 
  \end{figure*}

We use transactional data from instacart kaggle challenge to train all our models (sample data \ref{fig:sampledata}). From 
sample data we can see that data contains transactional details including
order id, add to cart order, date of transaction, aisle id and department id for each consumer-item transaction.
As described in Table \ref{tab:datasplit}, we utilize 1 year data for each consumer-item combination, 
which then gets split into train, validation, test1 and test2 as per our validation strategy. 
We generate consumer-item-week level data with purchase/ non purchase being the target,
and use this data to train all our models.

  \begin{figure}[t]
    \centering 
    \caption{Most reordered Items across Departments} 
    \includegraphics[width=2.25in]{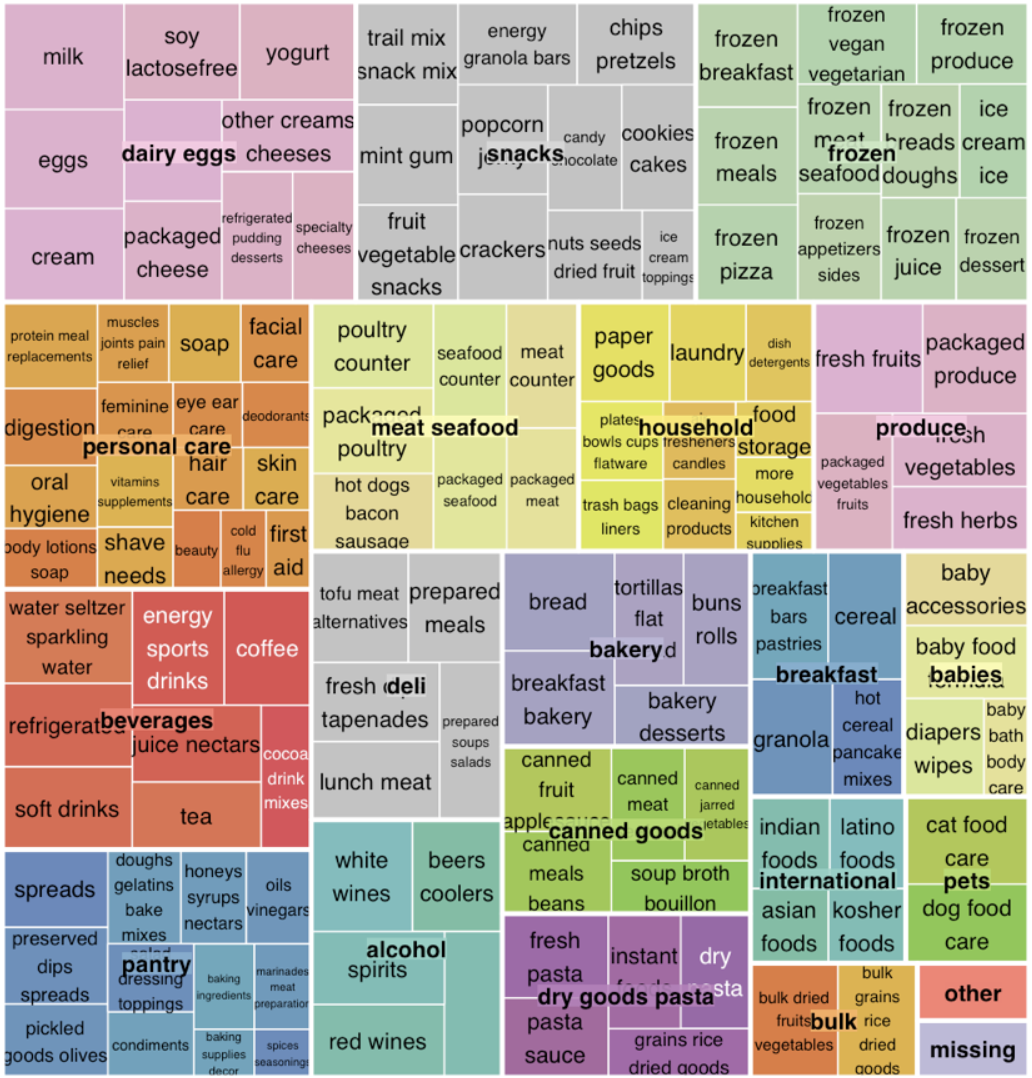} 
    \label{fig:items} 
  \end{figure}

  \begin{figure}[t]
    \centering 
    \caption{Density of consumers Vs. Basket Size} 
    \includegraphics[width=3in, height = 2in]{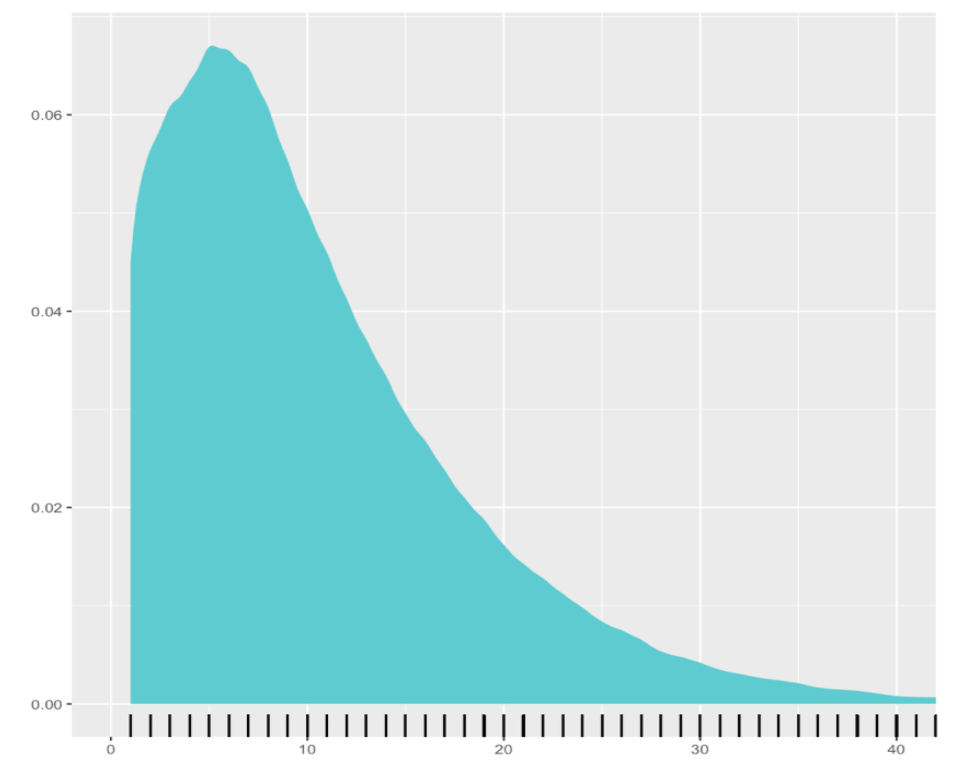} 
    \\ {\scriptsize \bf Basket size between 5-8 has maximum consumer density}
    \label{fig:basket} 
  \end{figure}

  \begin{figure}[t]
    \centering 
    \caption{Reorder probability Vs. Add to cart order} 
    \includegraphics[width=3in , height = 2in]{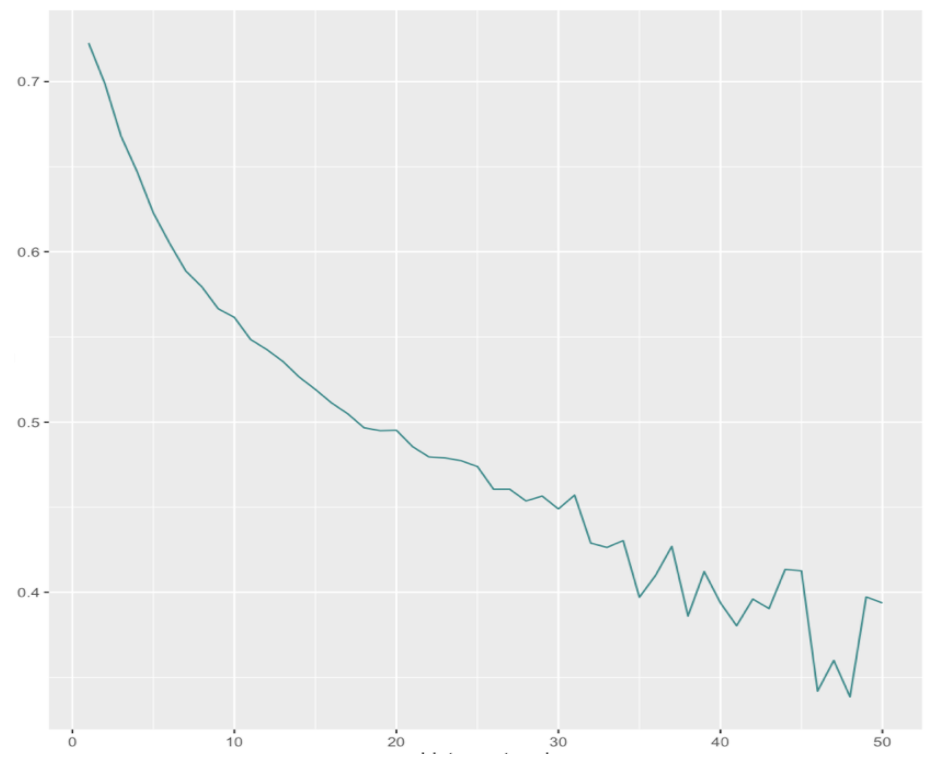} 
    \label{fig:addtocart} 
    \\ {\scriptsize \bf Probability of reordering decreases as the order of add to cart increases}
  \end{figure}

\begin{center}
\begin{table*}[!t]
\caption{BCELoss of Test2 for 12 Trials of Deep Learning Models} 
\centering
\resizebox{\textwidth}{!}{\begin{tabular}{|r|l|r|r|r|r|r|r|r|}
  \hline
 {\bf Trial} & {\bf Optimizer} & {\bf Scheduler} & {\bf SWA} & {\bf Parameter Avg} &  {\bf MLP} & {\bf LSTM} 
 &  {\bf TCN} & {\bf TCN-LSTM} \\ [0.5ex] 
  \hline\hline
1 & RMSprop & ReduceLROnPlateau & True & False &  {\bf 0.0276} & 0.0306 & {\bf 0.0249} & 0.0307 \\ 
2 & RMSprop & CyclicLR & True & False &  0.0708 & {\bf 0.0269} & {\bf 0.0269} & 0.0348 \\ 
3 & Adam & ReduceLROnPlateau & True & False &  0.0295 & 0.0303 & 0.0667 & 0.0337 \\ 
4 & RMSprop & ReduceLROnPlateau & False & False &  0.0297 & {\bf 0.0275} & 0.0364 & 0.0759 \\ 
5 & RMSprop & CyclicLR & False & False &  {\bf 0.0250} & 0.0306 & 0.0600 & {\bf 0.0286} \\ 
6 & Adam & ReduceLROnPlateau & False & False&  0.0360 & {\bf 0.0302} & 0.0590 & 0.0309 \\ 
7 & RMSprop & ReduceLROnPlateau & False & True &  0.0293 & 0.0432 & 0.0453 & 0.0381 \\ 
8 & RMSprop & CyclicLR& False & True &  {\bf 0.0245} & 0.0378 & 0.0569 & {\bf 0.0262} \\ 
9 & Adam & ReduceLROnPlateau & False & True & 0.0700 & 0.0491 & 0.0610 & 0.0382 \\ 
10 & RMSprop & ReduceLROnPlateau & True & True & 0.0356 & 0.0364 & {\bf 0.0238} & 0.0309 \\ 
11 & RMSprop & CyclicLR & True & True &  0.0420 & 0.0377 & 0.0284 & {\bf 0.0269} \\ 
12 & Adam  & ReduceLROnPlateau & True & True&  0.0321 & 0.0306 & 0.0547 & 0.0305 \\ [1ex] 
   \hline
\end{tabular}}
\label{tab:dlmodels}
\end{table*} 
\end{center}

\begin{table}[t]
\caption{BCELoss of Test2 for 6 best Trials of ML Models}
\vspace{0.1 in}
\centering
\resizebox{3.3in}{!}
{%
\begin{tabular}{|c|c|c|c|c|}
\hline
{\bf Trial} & {\bf HyperParameter} & {\bf Xgboost} & {\bf RandomForest} \\  
\hline\hline
1  		&  HyperOpt &  {\bf 0.0332} &  0.0526   \\ 
2	  		&  HyperOpt &  0.0364 &  0.0479   \\ 
3  		&  HyperOpt &  0.0347 &  {\bf 0.0416}  \\ 
4	  		&  HyperOpt &  0.0364 &  {\bf 0.0449}  \\ 
5	  		&  HyperOpt &  {\bf 0.0335} &  {\bf 0.0459}  \\ 
6	  		&  HyperOpt &  {\bf 0.0339} &  0.0578  \\ 
\hline
\end{tabular}
}
\label{tab:mlmodels}
\end{table}

\begin{table}[t]
\caption{ BCELoss mean of top 3 trials across data splits}
\vspace{0.1 in}
\centering
\resizebox{3.3in}{!}
{%
\begin{tabular}{|c|c|c|c|c|}
\hline
{\bf Model Type} & {\bf Val BCELoss} & {\bf Test1 BCELoss} & {\bf Test2 BCELoss} \\ 
\hline\hline 
MLP	  		&  0.0405 &  0.0289 &  0.0256  \\ \hline
LSTM  		&  0.0373 &  0.0293 &  0.0282 \\ \hline
{\bf TCN}			&  {\bf 0.0368}  &  {\bf 0.0292} &  {\bf 0.0251}  \\ \hline
TCNLSTM 	& 0.0368  & 0.0304	& 0.0273	 \\ \hline
Xgboost 	& 0.0352 & 0.0318	& 0.0335	\\ \hline
RandomForest & 0.0437 & 0.0389	& 0.0441	\\ \hline
\end{tabular}
}
\label{tab:training}
\end{table}

\begin{table}[t]
\caption{ Stacked Generalization Results}
\vspace{0.1 in}
\centering
\resizebox{3.3in}{!}
{%
\begin{tabular}{|c|c|c|c|c|c|}
\hline
{\bf Model Type} & {\bf K Value} & {\bf Val BCELoss} & {\bf Test1 BCELoss} & {\bf Test2 BCELoss} \\ 
\hline\hline 
{\bf Weighted K Best}	  &  {\bf 3}  &  {\bf 0.0386} &  {\bf 0.0278} &  {\bf 0.0242}  \\ \hline
Weighted K Best	  		&  5  &  0.0373 &  0.0282 &  0.0245  \\ \hline
Weighted K Best	  		 &  10 &  0.0397 &  0.0290 &  0.0258  \\ \hline
Weighted K Best	  		 &  15 &  0.0389 &  0.0296 &  0.0272  \\ \hline
Weighted K Best	  		&  25  &  0.0394 &  0.0316 &  0.0287  \\ \hline
\end{tabular}
}
\label{tab:stacking}
\end{table}

\begin{table}[hbt!]
\caption{Final Accuracy post F\textsubscript{1}-Maximization}
\vspace{0.1 in}
\centering
\resizebox{2.5in}{!}
{%
\begin{tabular}{|c|c|c|c|}
\hline
{\bf Data Split} & {\bf Precision} & {\bf Recall} & {\bf F\textsubscript{1}-Score} \\ 
\hline\hline 
Validation	  	 &  0.3401 &  0.4981 &  0.4042  \\ \hline
Test1	  		 &  0.3323 &  0.5103 &  0.4024  \\ \hline
Test2	  		 & 0.3506 &  0.4964 &  0.4109 \\ \hline
\end{tabular}
}
\label{tab:Fscore}
\end{table}

  \begin{figure}[hbt!]
    \centering 
    \caption{Probability Distributions} 
    \includegraphics[width=2in]{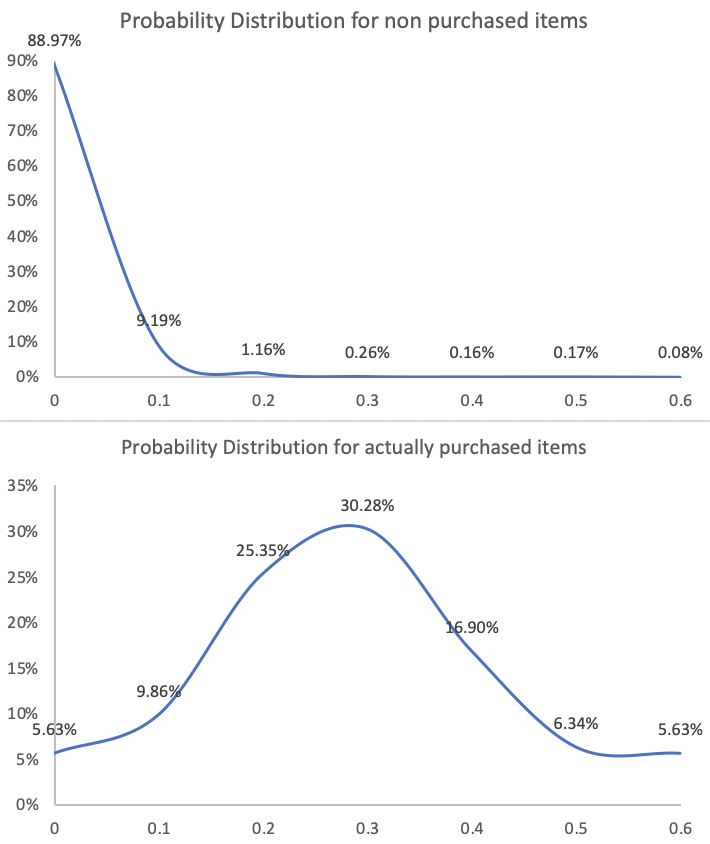} 
    \\ {\scriptsize \bf High density probability zone for the non purchased cases lies between 0 and 
    0.1, whereas for the purchased cases it lies between 0.25 and 0.35}
    \label{fig:probdensity} 
  \end{figure}
\subsection{Experiment Setups}
We start with exploratary data analysis, looking at the data from various cuts. We
study the variations of different features with our target (purchase/ non purchase). Some of our studies are
density of consumers versus basket size (Figure~\ref{fig:basket}), reorder visualization of items 
across departments (Figure \ref{fig:items}), variation of reorder probability with add to cart order (Figure \ref{fig:addtocart}),
order probability variations at different temporal cuts like week, month and quarter, transactional metrics 
like total orders, total reorders, recency, gap between orders, at both consumer and item levels.
We then perform multiple experiments with the above mentioned features and different hyperparameter configurations to land at reasonable 
hyperparameters to perform final experiments and present our results.

\subsection{Results and Observations}
Tables \ref{tab:dlmodels} and \ref{tab:mlmodels} show the experimental results obtained across models with different 
hyperparameter configurations. Table \ref{tab:dlmodels} contains the Deep Learning Experiment setup results
and Table \ref{tab:mlmodels} has Machine Learning model results. From model performance perspective, it is observed that 
Temporal Convolution Network (TCN) has least average BCELoss of 0.0251, approximately 2\% better than the second best model
which is Multi Layer Perceptron (MLP) having average BCELoss of 0.0256. Table \ref{tab:training} presents
the comparative analysis of average scores across all models. Also, we observe in Table \ref{tab:training} 
that Deep Learning models out perform Machine Learning models including Xgboost and RandomForest in terms of accuracy.
Average BCELoss of Deep Learning model is approximately about 0.0266 , whereas for Machine Learning models its 
approximately about 0.0388. From hyperparameter configuration perspective, we observe that RMSprop and CyclicLR emerged as the 
winners as Optimizer and Scheduler respectively (from Table \ref{tab:dlmodels}). 7 out of 12 times, the 
combination of RMSprop and CyclicLR (out of 3 possible combinations) generate the best result.

We also present the effectiveness of combining submodel predictions and F\textsubscript{1}-Maximization. 
Table \ref{tab:stacking} outlines the results of stacking at different values of K for Weighted K-Best stacker model. 
We realise the best accuracy or least BCELoss of 0.0242 at K = 3.
To analyse our probability values post stacking, we plot the probability distributions for both labels of the target,
as can be seen in Figure \ref{fig:probdensity}.
Finally we apply F\textsubscript{1}-Maximization over stacked probability values so as to generate 
purchase predictions. F\textsubscript{1}-Score Optimizer helps strike
balance between Precision and Recall \cite{buckland1994relationship}. Post F\textsubscript{1}-Maximization we 
observe that Precision, Recall and F\textsubscript{1}-Score are close enough for all data splits, as can be seen 
in Table \ref{tab:Fscore}. F\textsubscript{1}-Score of our model over unseen data (test2) is 0.4109
(Table \ref{tab:Fscore}).
\subsection{Industrial Applications}
The Next Logical Purchase framework has multiple applications in retail/e-retail industry. Some of them 
include:
\begin{itemize}
\item {\bf Personalized Marketing:} With prior knowledge of next logical purchase, accurate item recommendations and 
optimal offer rollouts can be made at consumer level. This will enable a seamless and delightful consumer 
shopping experience.
\item {\bf Inventory Planning:} Consumer preference model can also be used in better short term inventory planning (2-4 weeks),
which is largely dependant over what consumer is going to purchase in the near future.
\item {\bf Assortment Planning:} In retail stores, consumer choice study can be used to optimize the store 
layout with right product placement over right shelf.
\end{itemize}

\section{Conclusion}
\label{sec:conclusion}
We have presented our study of the consumer purchase behaviour in the context of large scale e-retail. 
We have shown that careful feature
engineering when used in conjunction with Deep Neural Networks, can be used
to predict the next (multi-timestep) logical purchase of consumers with reasonably good accuracies.
While creating our models and features we have been cognizant of
the fact that many features might not be available when predictions
are being generated for future period. Hence we save model weights and use innovative transformations 
so that we do not necessarily have to remember complete data during forecast time,
thereby reducing the computation and memory requirements during forecast generation.

As per our initial expectations, Deep Neural Network models outperform the ML models like Xgboost and RandomForest.
Sequence to Sequence architectures seems to be sound choice for tackling our problem, and our results and 
observations are inline with this thought process. Model generalization and robustness
is attained using stacked generalization. As per our expectation we realize gain in accuracy post both stacking
and F\textsubscript{1}-Maximization.

At the same time we understand that computation strategy is a key aspect in modelling
millions of consumers, and we intend to further explore this aspect by building 
Transfer Learning framework \cite{yosinski2014transferable}. We are also working to 
further improve our Sequence to Sequence neural network architectures to improve accuracy
and decrease computation time.

\section{Acknowledgements}
The authors would like to thank Yadunath Gupta for his contributions towards Neural network architecture improvements. 
Siddharth Shahi, Deepinder Singh Dhingra and Ankit Jain for helpful discussions and insights over the methodology
and implementation.
\bibliography{paper}
\end{document}